# Deep Learning for Predicting Asset Returns

Guanhao Feng[*]

*College of Business*

*City University of Hong Kong*

Jingyu He[†]

*Booth School of Business*

*University of Chicago*

Nicholas G. Polson [‡]

*Booth School of Business*

*University of Chicago*

April 25, 2018

## Abstract

Deep learning searches for nonlinear factors for predicting asset returns. Predictability is achieved via multiple layers of composite factors as opposed to additive ones. Viewed in this way, asset pricing studies can be revisited using multi-layer deep learners, such as rectified linear units (ReLU) or long-short-term-memory (LSTM) for time-series effects. State-of-the-art algorithms including stochastic gradient descent (SGD), `TensorFlow` and dropout design provide implementation and efficient factor exploration. To illustrate our methodology, we revisit the equity market risk premium dataset of Welch and Goyal (2008). We find the existence of nonlinear factors which explain predictability of returns, in particular at the extremes of the characteristic space. Finally, we conclude with directions for future research.

---

[*]Address: 83 Tat Chee Avenue, Kowloon Tong, Hong Kong. E-mail address: `gavin.feng@cityu.edu.hk`.
[†]Address: 5807 S Woodlawn Avenue, Chicago, IL 60637, USA. E-mail address: `jingyu.he@chicagobooth.edu`.
[‡]Address: 5807 S Woodlawn Avenue, Chicago, IL 60637, USA. E-mail address: `ngp@chicagobooth.edu`.

# 1 Introduction

Deep learning searches for nonlinear factors to predict asset returns via a composition of factor-based characteristics. Predicting asset returns is important to empirical finance and factor models play a central role, for example, see Rosenberg et al. (1976) and Fama and French (1993). Cross-sectional time series predictability is studied using predictive regressions, see Kandel and Stambaugh (1996), Barberis (2000) and Welch and Goyal (2008). We build on this line of research, by incorporating deep learning with hierarchical layers of nonlinear factors to perform out-of-sample prediction. Deep learning factors provide improvements at the extremes of the characteristic space when explaining empirical asset returns. They provide an alternative to dynamic factor modeling, see Lopes and Carvalho (2007), Carvalho et al. (2011) and Carvalho et al. (2017).

While the use of (artificial) neural networks is not novel in economics and finance, see Gallant and White (1988a,b); Hornik et al. (1989); Gallant and White (1992); Kuan and White (1994); Hutchinson et al. (1994); Lo (1994); Qi (1999); Jones (2006); Sirignano et al. (2016) and Heaton et al. (2017), deep learning is new to characteristic-based asset pricing. Deep learning is capable of extracting nonlinear factors and provides a powerful alternative to feature selection and shrinkage methods. Feng et al. (2017). Recent work of Kozak et al. (2017); Gu and Xiu (2018) shows the promise of machine learning based predictors in empirical finance, including traditional regularization methods and trees and. Feng et al. (2018) predicts cross-sectional returns with deep learning in a portfolio context. Deep networks, as opposed to shallow ones, can achieve out-of-sample performance gains versus linear additive models, while avoiding the curse of dimensionality, for example, see Poggio et al. (2017).

To predict the equity premium with a large set of economic variables, Welch and Goyal (2008) leads to out-of-sample performance which is hard to outperform historical means. Improved methodologies have been suggested by Campbell and Thompson (2007), Rapach et al. (2010) and Harvey et al. (2016) and, more recently, Feng et al. (2017) and Kozak et al. (2017). The latter relies on shrinking the cross-sectional select which economic variables are of importance, rather than deep learning which extracts nonlinear factors from the full characterisitc space with a goal of improved predictive performance.

The rest of our paper is organized as follows. Section 1.1 discusses deep learning in financial economics. Section 2 constructs our deep learning architectures for applications in forecasting the equity premium. Section 3 shows simulation results. Section 4 reports our results on predicting stock returns, mimicking the analysis of <span>Welch and Goyal</span> (<span>2008</span>). Finally, appendices contain details on SGD and LSTM models as well as a comprehensive set of results on our empirical study.

## 1.1 Deep Learning Econometrics

Deep learning is a form of supervised learning for predicting an output variable, $Y$, via predictors, $X$. Deep learning comprises of a series of $L$ non-linear transformations applied to the input space $X$. Each of the $L$ transformations is referred to as a layer, where the original input is $X$, the output of the first transformation is the first layer, and so on, with the output $\hat{Y}$ as the $(L+1)$-th layer. We use $l \in \{1, \ldots, L\}$ to index the layers from $1$ to $L$, which are called hidden layers. The number of layers $L$ represents the depth of the architecture.

Specifically, a deep neural network can be described as follows. Let $f_1, \ldots, f_L$ be given univariate activation functions for each of the $L$ layers. Activation functions are non-linear transformations of weighted data. Commonly used activation functions are sigmoidal (e.g., $1/(1 + \exp(-x))$ or $\tanh(x)$), heaviside gate functions (e.g., $\mathbb{I}(x > 0)$), or rectified linear units (ReLU) $\max\{x, 0\}$. We let $Z^{(l)}$ denote the $l$-th layer which is a vector with same length as number of neurons in that layer, and so $X = Z^{(0)}$. The explicit structure of a deep prediction rule is then a composition of univariate semi-affine functions,

$$
\begin{aligned}
F^{W,b} &= F_1^{W^{(1)},b^{(1)}} \circ \cdots \circ F_L^{W^{(L)},b^{(L)}} \\
F_l^{W^{(l)},b^{(l)}} &:= f_l(W^{(l)}Z^{(l)} + b^{(l)}) = f_l\left(\sum_{i=1}^{N_l} W_i^{(l)} Z_i^{(i)} + b_i^{(l)}\right), \quad \forall 1 \leq l \leq L
\end{aligned}
\tag{1}
$$

where $N_l$ is the number of neurons or width of the architecture at layer $l$. $W^{(l)}$ are real weight matrices and $b^{(l)}$ are threshold or activation level which contribute to the output of a hidden unit, allowing the activation function to be shifted left or right. One noticeable property is that the weights $W_l \in \mathbb{R}^{N_l \times N_{l-1}}$ are matrices. In an econometric perspective, deep learner models constitute a particular class of nonlinear neural network predictors. $F_l$ denotes the $l$-th hidden layer. As in traditional financial modeling we can view $F^{(l)}$ as latent factors. The main difference is that we will use a *composition* of factors versus a traditional additive structure. Moreover, the hidden factors $F^{(l)}$ will be extracted from the algorithm.

## 2    Deep Learning for Characteristic Based Asset Pricing

Let $R_{t+1} \in \mathbb{R}^{T \times 1}$ be a vector of asset returns, $X_t \in \mathbb{R}^{T \times p}$ a high dimensional set of predictor variables. Deep learning is a data reduction scheme that uses $L$ layers of "hidden" factors, which can be highly nonlinear. The factors are extracted from data set with the dual goal of good out-of-sample prediction and in-sample model fit. Mean squared prediction error (MSPE) is a common metric for out-of-sample predictor performance. From a finance viewpoint, we have a hierarchical model of the form

$$
\begin{aligned}
R_{t+1} &= \alpha + \beta X_t + \beta_f F_t + \epsilon_{t+1} \\
F_t &= F^{W,b}(X_t) \\
F^{W,b} &:= f_1^{W_1,b_1} \circ \cdots \circ f_L^{W_L,b_L} \\
f^{W_l,b_l}(Z) &:= f_l(W_l Z + b_l), \quad \forall 1 \le l \le L
\end{aligned}
\tag{2}
$$

where $F : \mathbb{R}^{T \times p} \to \mathbb{R}^{T \times 1}$ is a multivariate data reduction map represented as a deep learner. The network parameters $(W, b)$ are weights and offsets to be *trained*. Here $\epsilon_t$ are the usual idiosyncratic pricing errors. The major difference between DL and traditional factor models are the useage of compositions of factors rather shallow additive models. $F$ is constructed as a composition of univariate semi-affine functions and a common choice for activation function is $f_l(x) = \max(x, 0) :=$ ReLU$(x)$, the so-called rectified linear unit. This leads to Deep ReLU networks which have been popular in applications from image processing to game intelligence.

Traditionally, researchers estimate factors $F_t$ and then learn coefficients $\alpha, \beta$ by regression with a two-step procedure. Here $R_{t+1}$ is a linear additive combination of input variables $X_t$ and latent factors $F_t$,

$$
R_{t+1} = \alpha + \beta X_t + \beta_f F_t
$$

Deep learning will estimate coefficient $\alpha, \beta$ and latent factors, $F_t$, jointly. Figure 1 illustrates this with green circles on left side as input predictors $X_t$ for example, dividends, earnings, inflation or other economic variables. The key advantage is non-linearity and simultaneous factor estima-tion. The rightmost red circle is asset return $R_{t+1}$ to predict. The purple circles are fully connected neurons in hidden layers. The yellow circles are the last hidden layer in equation 2, but are different from the first hidden layer as they are composed of latent factors $F_t$ generated by previous hidden layers and a copy of the original input $X_t$.

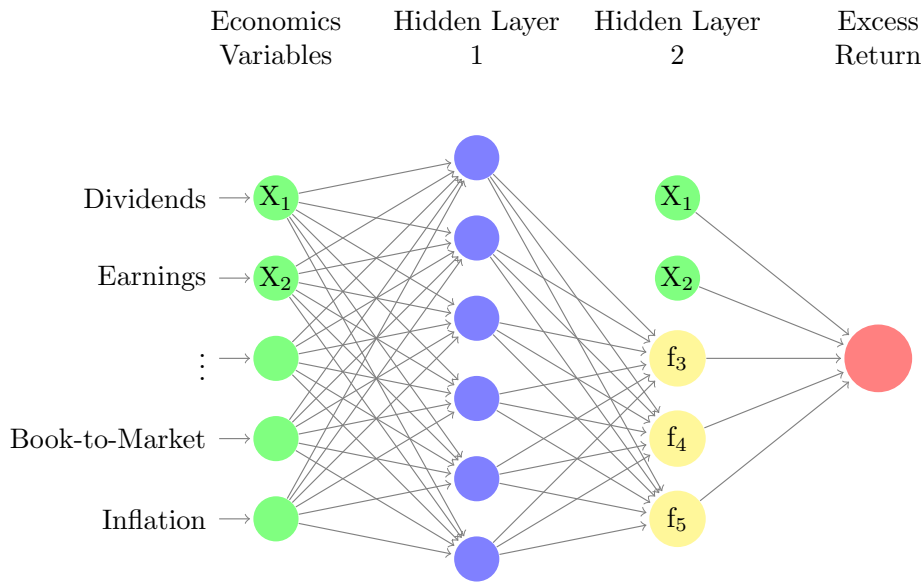

Figure 1: Deep Learning architecture to estimate dynamic factor model

To *train* a model, we need a loss function to minimize, which is typically mean squared error of the in-sample fit of $\hat{R}_{t+1}$.

$$L = \frac{1}{T}\sum_{t=1}^{T}\left(R_{t+1} - \hat{R}_{t+1}\right)^{\intercal}\left(R_{t+1} - \hat{R}_{t+1}\right) + \lambda\phi\left(\beta, W, b\right), \tag{3}$$

where $\phi\left(\beta, W, b\right)$ is a regularization penalty to induce predictor selection and avoid model over-fitting and $\lambda$ controls the amount of regulations. The regressor parameters $\alpha, \beta, \beta_f$ and factors $F_t$ *jointly* minimize loss function using stochastic gradient descent (SGD) algorithm in `TensorFlow`.

Another notable difference is that there is no stochastic error in the factor construction. Kandel and Stambaugh (1996) discuss the difficulty of the model estimation and prediction in the presence of parameter uncertainty. They propose a Bayesian framework to add a regularization prior on the predictor existence and strength. Their model still cannot deal with an ultra high-dimensional $x_t$ as well as its nonlinear signals.

## 3 Application

### 3.1 Simulation Study

To illustrate the possible gains available in deep learners for predictions, we provide a simulation study to compare its prediction performance versus traditional machine learning tools such as linear regression, Lasso, partial least squares and elastic net. As data generating processes, we use one layer and two layers dynamic latent factor model.

$$\mathbf{R}_{t+1} = \alpha + \beta \mathbf{F}_t + \epsilon_{t+1}$$

$$\mathbf{F}_t = f_l(W \mathbf{X}_t + b)$$

Similarly, the two layers data generating process is

$$\mathbf{R}_{t+1} = \alpha + \beta \mathbf{F}_{2,t} + \epsilon_{t+1}$$

$$\mathbf{F}_{1,t} = f_l(W^{(1)} \mathbf{X}_t + b^{(1)})$$

$$\mathbf{F}_{2,t} = f_l(W^{(2)} \mathbf{F}_{1,t} + b^{(2)})$$

The one hidden layer data generating process is described as follows. Suppose the length of time series is $T$. $\mathbf{X}_t$ is a $T \times p$ matrix drawn from i.i.d. standard normal distribution. $\alpha$, $\beta$, $W$ and $b$ are coefficients also drawn from i.i.d. standard normal distribution. We control in-sample $R_{IS}^2$ as a measure of signal level, draw $\epsilon_{t+1}$ from i.i.d. $N(0, \sigma^2)$ where $\sigma^2$ are solved from equation

$$R_{IS}^2 = \frac{\text{var}(\alpha + \beta \mathbf{Z}_t)}{\text{var}(\alpha + \beta \mathbf{Z}_t) + \sigma^2}.$$

where $W^{(1)}$, $W^{(2)}$, $b^{(1)}$, $b^{(2)}$ are coefficients drawn from i.i.d. standard normal. $\mathbf{X}_t$ is a $T \times p$ matrix drawn from i.i.d. standard normal.

In the data generating process above, we generate $R^2$ in the same way and regressors $\mathbf{X}_t$, output $\mathbf{y}_{t+1}$ and latent factors $\mathbf{F}_t$. Three different deep learning architectures are compared: three hidden layer deep learning model with 64 neurons in the first hidden layer, 32 neurons in the second hidden layer and 16 neurons in the third hidden layer (DL-64-32-16). We add $\mathbf{X}_t$ to the last layer (as shown in Figure 1), so there are actually $p + 16$ neurons in the third layer. Similarly we create two hidden layer deep learning model with 32 and 16 neurons in each layer (DL-32-16) and one 16 neurons hidden layer model (DL-16). To illustrate the success of deep learning models, we also compare with other frequently used approach such as ordinary least squares (OLS), partial least squares (PLS), Lasso, elastic-net and "oracle OLS" which regress $\mathbf{y}_t$ on latent factors $\mathbf{F}_t$ directly. Oracle OLS is not possible to implement as we need to learn latent factors $\mathbf{F}_t$ from data. Shrinkage parameters of Lasso and elastic net are selected by cross validation. For comparation, we use linear regression, Lasso, Ridge, Elastic Net, PCA and PLS regressions. Simulation studies reveal that neural network outperforms all other methods.

Table 1 provides results of one layer data generating process. Table 2 shows results of two layers data generating process. It is straightforward to see that oracle OLS achieves smallest mean squared prediction error (MSPE). The three deep learning models (DL 64-32-16, DL 32-16 and DL 16) are very similar to each other in all circumstances and all of them are much better than other methods except oracle OLS as expected. Deep learning's advantage over other approaches diminishes a little bit as $R^2$ increases (noise level decreases). Table 1 shows the same pattern as Table 2, our approach is the closest to oracle OLS.

| K | T | $R^2$ | MSPE | | | | | | | |
|---|---|---|---|---|---|---|---|---|---|---|
| | | | Oracle | OLS | PLS | Lasso | ElasNet | DL 64-32-16 | DL 32-16 | DL 16 |
| 5 | 500 | 0.05 | 97.19 | 160.06 | 148.36 | 155.91 | 141.74 | 122.18 | 122.32 | 122.91 |
| 5 | 500 | 0.25 | 22.46 | 41.46 | 37.76 | 39.43 | 34.4 | 31.74 | 31.87 | 32.77 |
| 5 | 500 | 0.5 | 4.32 | 9.42 | 8.53 | 8.65 | 7.5 | 7.4 | 7.33 | 7.78 |
| 5 | 500 | 0.75 | 1.93 | 6.25 | 5.74 | 5.82 | 5.31 | 5.17 | 5.17 | 5.47 |
| 25 | 500 | 0.05 | 588.16 | 874.51 | 800.94 | 864.13 | 813.82 | 665.27 | 660.89 | 650.71 |
| 25 | 500 | 0.25 | 93.49 | 159.21 | 145.91 | 154.74 | 140.71 | 123.59 | 123.49 | 123.63 |
| 25 | 500 | 0.5 | 22.56 | 44.51 | 40.94 | 42.75 | 38.42 | 35.43 | 35.27 | 36.15 |
| 25 | 500 | 0.75 | 12.85 | 36.41 | 32.42 | 34.9 | 31.58 | 29.61 | 29.48 | 30.54 |

Table 1: One layer data generating process. $P$ is dimension of $X_t$, $K$ is dimension of latent factors $Z_t$, $T$ is length of time series. $K_P LS$ is number of factors used in principal regression. $R^2$ is R squared of DGP, $R_{t+1} = \alpha + \beta F_t + \epsilon_{t+1}$. The oracle regresses $R_t$ on $F_t$, all other methods regress $R_t$ on $x_t$. Dropout = 0.5

| | | | | | MSPE | | | | | |
|---|---|---|---|---|---|---|---|---|---|---|
| P | $K_1$ | $K_2$ | T | $R^2$ | Oracle | OLS | PLS | Lasso | Elastic-Net | DL 64-32-16 |
| 100 | 25 | 25 | 500 | 0.05 | 1130.05 | 1401.06 | 1302.05 | 1390.49 | 1329.43 | 1088.67 |
| 100 | 25 | 25 | 500 | 0.25 | 121.57 | 160.04 | 143.22 | 155.72 | 140.31 | 117.92 |
| 100 | 25 | 25 | 500 | 0.5 | 22.74 | 38.48 | 35.98 | 37.19 | 34.04 | 32.17 |
| 100 | 25 | 25 | 500 | 0.75 | 8.61 | 19.57 | 17.64 | 18.67 | 16.95 | 15.62 |
| 100 | 25 | 5 | 500 | 0.05 | 146.81 | 244.75 | 223.29 | 239.48 | 219.26 | 183.25 |
| 100 | 25 | 5 | 500 | 0.25 | 15.35 | 30.62 | 27.58 | 28.94 | 24.75 | 21.61 |
| 100 | 25 | 5 | 500 | 0.5 | 4.19 | 9.25 | 8.50 | 8.65 | 7.61 | 7.03 |
| 100 | 25 | 5 | 500 | 0.75 | 1.14 | 3.85 | 3.54 | 3.50 | 3.18 | 3.01 |

Table 2: Two layers data generating process. $P$ is dimension of $X_t$, $K_1$ and $K_2$ are dimensions of latent factors $Z_t^1$ and $Z_t^2$ respectively. $T$ is length of time series. $K_P LS$ is number of factors used in principal regression. $R^2$ is R squared of DGP, $R_{t+1} = \alpha + \beta F_t + \epsilon_{t+1}$. Oracle regresses $R_t$ on $F_t$, all other methods regress $R_t$ on $x_t$.

## 4    Predict Asset Returns

**Data.** Predicting market excess returns is a challenge as usually predictors have difficulty in outperforming historical mean averages. Welch and Goyal (2008) explore the out-of-sample excess market return predictability of S&P 500 based on a large set of economic predictor variables.

Table 3 provides the variable descriptions. The sample frequency is monthly data, beginning in 1926 December to 2016 December, giving 90 years in total.

| Variable | Description |
|---|---|
| **d/p** | dividend price ratio |
| **d/y** | dividend yield ratio |
| **e/p** | earning price ratio |
| **d/e** | dividend payout ratio |
| **svar** | stock variance |
| **b/m** | book to market ratio |
| **ntis** | net issues |
| **tbl** | treasury bills rate |
| **ltr** | long term rate |
| **tms** | term spread |
| **dfy** | default spread |
| **dfr** | default return spread |
| **infl** | consumer price index |
| **cay** | consumption, wealth, income ratio |

Table 3: Description of all variables

**Predictive Regressions.** A predictive regression model takes the form

$$R_{t+1} = \alpha + \beta^{\mathsf{T}} X_t + \epsilon_{t+1}$$

$$X_t = A + B X_{t-1} + u_t.$$

Here, $R_{t+1}$ is the logarithm return on a market portfolio S&P 500. $X_t$ is a $p \times 1$ predictor from the lag period and follows a VAR(1) model.

Our deep learning dynamic factor model builds on this by assuming

$$R_{t+1} = \alpha + \beta X_t + \beta_f F_t + \epsilon_{t+1}$$

$$F_t = F^{W,b}(X_t)$$

Deep learning structures are flexible enough that we need to specify number of hidden layers and number of neurons in each layer. We present result of a variety of deep learning structures (DL1 to DL4) and compare with a range of frequently used methods such as ordinary least squares (OLS), ridge regression (Ridge), partial least squares (PLS), Lasso and elastic-net. For example, for model DL1, 32-16-8 means a three layers deep network with 32, 16 and 8 neurons in the three hidden layers respectively. The shrinkage level is a critical tuning parameter for Lasso and elastic net, which is selected by cross validation.

| Method | Description |
|--------|-------------|
| DL1 | 32-16-8 |
| DL2 | 16-8-4 |
| DL3 | 16-8 |
| DL4 | 16 |
| OLS | ordinary least squares |
| Ridge | ridge regression |
| PLS | partial least squares |
| Lasso | Lasso |
| ElasNet | elastic net |

Table 4: Description of all methods for comparision

**Input variables.** We consider four different cases of input variables with growing dimensionality. In the first case, the input variables are the original 14 variables $X_t$ as shown in table 3. The second case has squared terms $(X_t, X_t^2)$ and 28 variables in total. In the third case we add an asymmetric term $(X_t, \mathbb{I}\{R_{t-1} > 0\} \times X_t)$ where $\mathbb{I}\{R_{t-1} > 0\}$ is an indicator whether the return of previous period is positive or not. In the last case, we use all variables above as $(X_t, X_t^2, \mathbb{I}\{R_{t-1} > 0\} \times X_t)$, 42 variables in total.

**Output variables.** We predict the excess logarithm return of S&P 500, which is the difference between logarithm return of S&P 500 and risk-free interest rate. Welch and Goyal (2008) points out that although it is not clear how to choose the periods over which a regression model is estimated and subsequently evaluated, it is important to have enough initial data to get a reliable regression estimate at the start of the evaluation period. We predict three kinds of returns: one, three and twelve months returns.

**Training set specifications.** We also explore two period specifications. The first one uses a fixed (50 years) moving window to estimate the model and predicts the following period. The second one uses a cumulative moving window which starts from 1926 December to the current month and predicts the next period. Therefore the second specification uses more and more data as the window moving with time.

**Forecast Evaluations.** To compare prediction results $\hat{R}_{t+1}$ and $R_{t+1}$, we use out-of-sample $R_{OS}^2$ as suggested by Campbell and Thompson (2007). The $R_{OS}^2$ statistics is akin to in-sample $R^2$ and is defined by

$$R_{OS}^2 = 1 - \frac{\sum_{k=P_0+1}^{P}(R_{t+k} - \hat{R}_{t+k})^2}{\sum_{t=P_0+1}^{P}(R_{t+k} - \bar{R}_{t+k})^2}. \tag{4}$$

where $\hat{R}_{t+k}$ is the prediction of a predictive model. $\bar{R}_{t+k}$ is forecast of historical mean. When $R_{OS}^2 > 0$, the predictive regression model has better mean squared prediction error (MSPE) than the benchmark average historical return. The most popular method for testing significant difference in MSPE is the Diebold and Mariano (2002) and West (1996) test. Even if there is evidence that $R_{OS}^2$ is statistically significant, $R_{OS}^2$ values are typically small for prediction models, but as Campbell and Thompson (2007) and Rapach et al. (2010) argue, even very small the $R_{OS}^2$ values can be economically meaningful in terms of portfolio returns. We find the same is true for our non-linear deep learning prediction rules. Figure 2 shows histogram of $R_{OS}^2$ for all methods and more comprehensive tables are provided in the appendix.

**Empirical Results.** Figure 2 shows MSPE and $R^2$ of 1-month predictions for moving window and cumulative windows. Comprehensive tables can be found in the appendix. Here we summarize the broad trends that merges. All other methods get negative $R^2_{OS}$, which is consistent with conclusions of Welch and Goyal (2008) that all predictive regression models cannot beat simple historical mean. However, most deep learning approaches get slightly positive $R^2_{OS}$ in all four different input cases, which indicates that they have a smaller mean squared error than historical mean. The results of predicting one month return using cumulative width moving window are shown in table 6, where deep learning methods achieve higher $R^2_{OS}$ than fixed width 600-month moving window because deep learning model can learn the nonlinear structures better with more training observations. Note that OLS has much lower negative $R^2_{OS}$ in table 6 because it cannot learn the complex trend well in a longer time horizon.

Figure 3 shows fitted latent factor of DL1 (32-16-8) model. We plot the 8 latent factors against the first 4 input variables **d/p**, **d/y**, **e/p** and **d/e**. It straightforward to see that deep learning model learns highly nonlinear factors from the data.

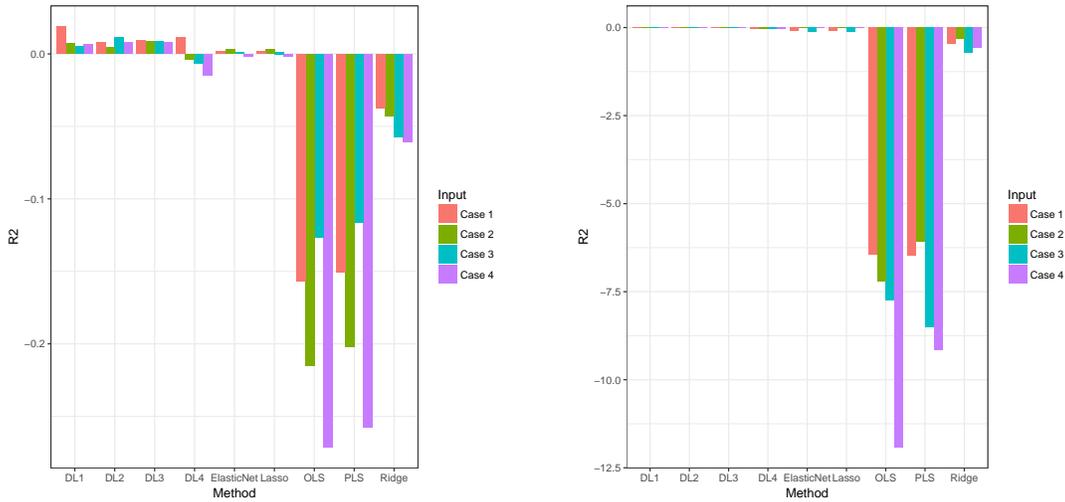

(a) $R^2$ of 1 month return predictions. Moving window.

(b) $R^2$ of 1 month return predictions. Cumulative window.

Figure 2: We generate 8 latent factors by fitting a three layer deep dynamic factor model with 32, 16 and 8 neurons in each layer. The plot shows the latent factors against first 4 variables **d/p**, **d/y**, **e/p** and **d/e**. Strong non-linearity exists.

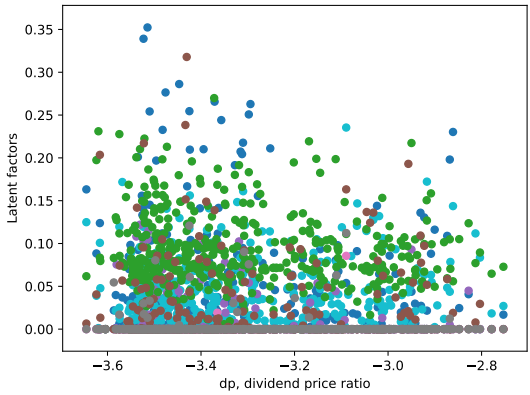

(a) **d/p**, dividend price ratio

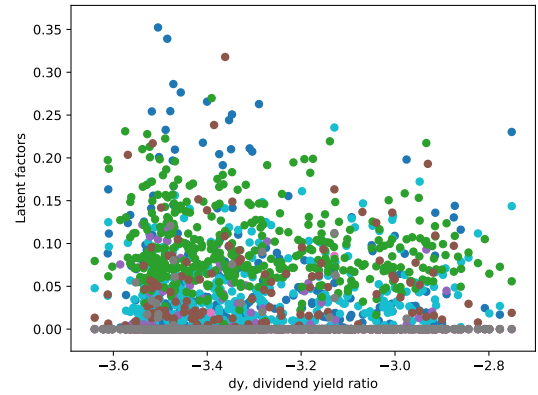

(b) **d/y**, dividend yield ratio

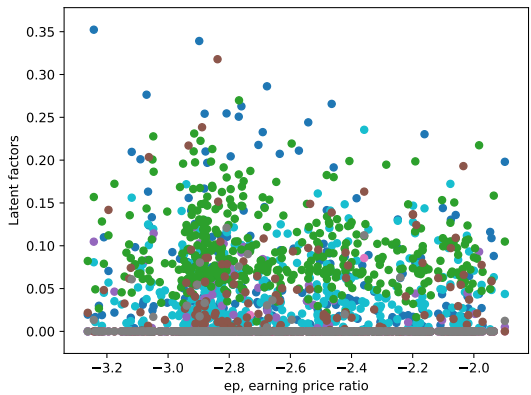

(c) **e/p**, earning price ratio

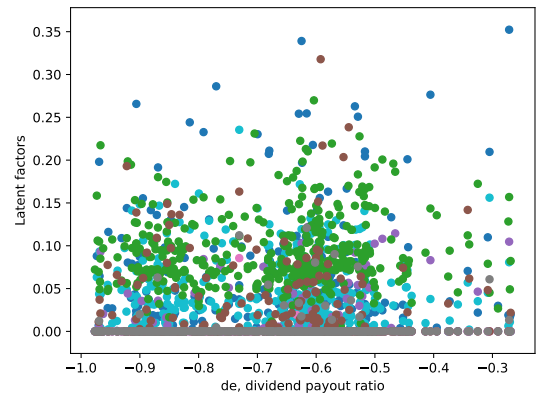

(d) **d/e**, dividend payout ratio

Figure 3: We generate 8 latent factors by fitting a three layer deep dynamic factor model with 32, 16 and 8 neurons in each layer, using input variables in case 1. The plot shows 8 latent factors against first 4 variables **d/p**, **d/y**, **e/p** and **d/e**. Each color indicates one latent factor.

**Comparison with Trees.** We do further analysis to show the difference between deep learning and tree based methods. Instead of prediction the logarithm return of S&P 500, we predict whether the market crash or not, where the definition of a crash is that monthly return is worse than $-10\%$. Therefore, the regression problem is converted to a classification problem.

Figure 4 shows the in-sample fit of deep learning and tree based methods using two economic variables dividend price ratio (dp) and inflation (infl). It's straightforward to see that Both CART and boosting always predict "no crash" no matter what value the input variables take. Deep learning does slightly better than random forest because it captures more tail behavior. However, none of the machine learning methods can achieve good in-sample fit since the input variables are way too noisy. Samuelson famously said "Economists have forecast nine of the last five recessions". Unlike pessimistic economic predictions, deep learning is (overly) optimistic and predict less market crashes than the truth, but more than random forest.

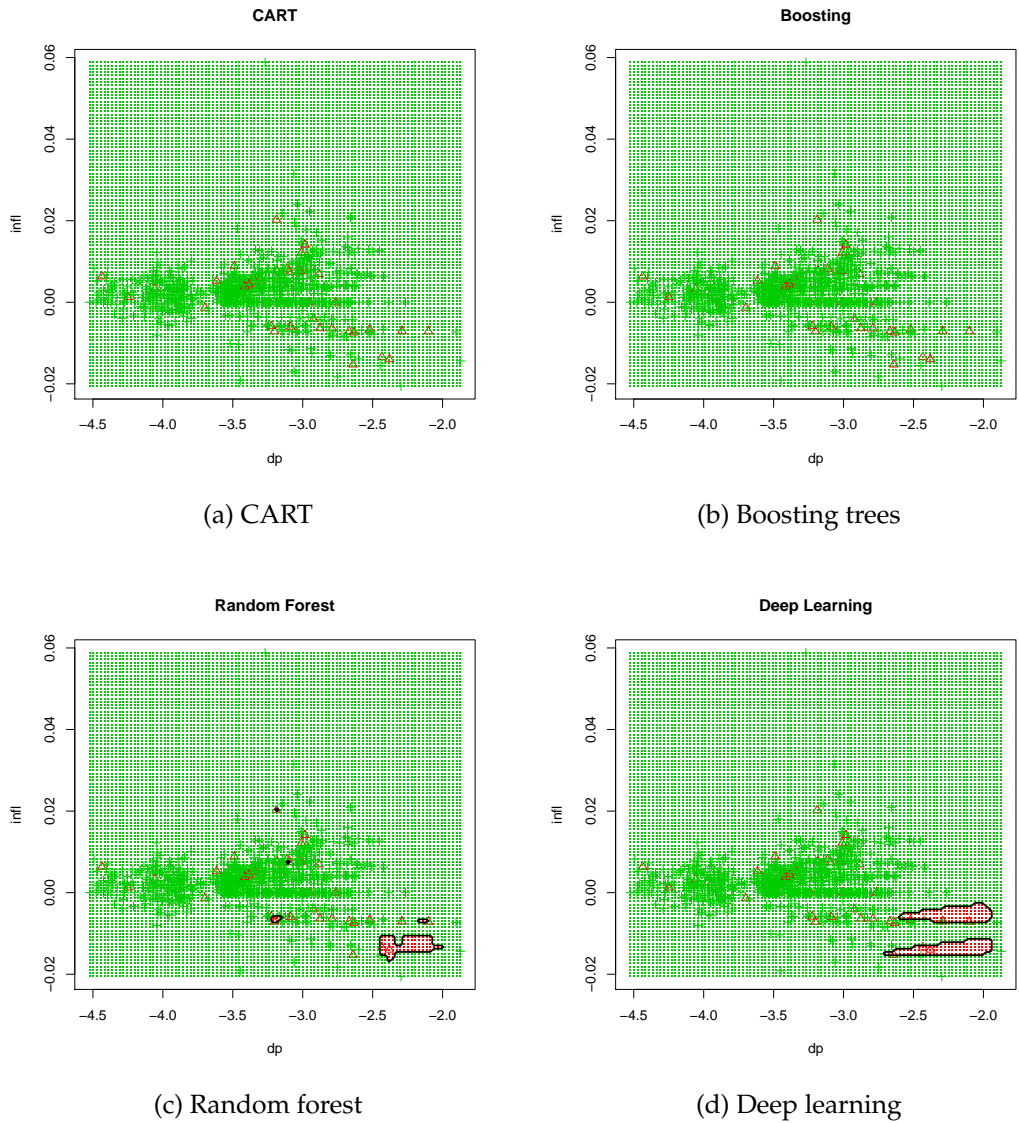

Figure 4: In-sample prediction and decision boundary of tree based methods and deep learning. Green cross indicates no market crash and red triangular indicates market crash where the monthly return is smaller than $-10\%$.

# 5 Discussion

Deep learning dynamic factor models are constructed for predicting asset returns. Both hidden factors and regression coefficients are jointly estimated by stochastic gradient descent. Deep learning is a very flexible class of machine learning tools for empirical analysis. By varying the number of hidden layers and the number of neurons within each layer, very flexible predictors can be training, and out-of-sample cross-validation provides a technique to avoid overfitting. Long-short-term-memory (LSTM) models are alternatives to traditional state space modeling.

Deep learning methods have some advantages and caveats. The key advantages are: (i) With `TensorFlow`, it is easy to implement deep learning architectures, (ii) Composite versus additive models, (iii) Hyperplanes versus cylinder sets. With the coefficient term $W$, deep learning model can rotate input variables and create cutoff hyperplanes. Hence better classification rules, (iv) Able to fit in-sample far more accurately.

The key caveats are (i) Model interpretability, (ii) Learn only correlation but not causation, (iii) Despite many gains from neural networks to detect and exploit interactions in empirical that are hard to identify using existing economic theory, they have several important limitations. In particular, perform causal inference from large datasets is hard when there are complex data interactions without taking assumptions for economic model specification. Due to the nesting of layers, statistical inference cannot always be applied to deep learning. Yet, deep learning provides a very fruitful linear of research particularly in empirical asset pricing studies.

# A Complete Results of Predicting Asset Returns

## A.1 1 month return prediction

| | Case 1 | | Case 2 | | Case 3 | | Case 4 | |
|---|---|---|---|---|---|---|---|---|
| | MSPE | $R^2_{OS}$ | MSPE | $R^2_{OS}$ | MSPE | $R^2_{OS}$ | MSPE | $R^2_{OS}$ |
| OLS | 0.00224 | -0.15680 | 0.00235 | -0.21513 | 0.00218 | -0.12647 | 0.00242 | -0.27127 |
| Ridge | 0.00201 | -0.03761 | 0.00202 | -0.04316 | 0.00205 | -0.05721 | 0.00202 | -0.06075 |
| PLS | 0.00223 | -0.15022 | 0.00233 | -0.20209 | 0.00216 | -0.11637 | 0.00240 | -0.25753 |
| Lasso | 0.00193 | 0.00165 | 0.00193 | 0.00322 | 0.00193 | 0.00138 | 0.00191 | -0.00164 |
| ElasticNet | 0.00193 | 0.00145 | 0.00193 | 0.00313 | 0.00193 | 0.00133 | 0.00191 | -0.00157 |
| DL1 | 0.00192 | 0.01864 | 0.00192 | 0.00695 | 0.00193 | 0.00517 | 0.00189 | 0.00665 |
| DL2 | 0.00192 | 0.00784 | 0.00193 | 0.00441 | 0.00192 | 0.01111 | 0.00189 | 0.00766 |
| DL3 | 0.00192 | 0.00941 | 0.00192 | 0.00885 | 0.00192 | 0.00851 | 0.00189 | 0.00761 |
| DL4 | 0.00195 | 0.01103 | 0.00194 | -0.00355 | 0.00195 | -0.00654 | 0.00193 | -0.01471 |

Table 5: Prediction 1 month logarithm return of S&P 500. The training data is a fixed length 600 month moving window.

| | Case 1 | | Case 2 | | Case 3 | | Case 4 | |
|---|---|---|---|---|---|---|---|---|
| | MSPE | $R^2_{OS}$ | MSPE | $R^2_{OS}$ | MSPE | $R^2_{OS}$ | MSPE | $R^2_{OS}$ |
| OLS | 0.01346 | -6.43907 | 0.01487 | -7.22048 | 0.01589 | -7.74735 | 0.02347 | -11.91466 |
| Ridge | 0.00263 | -0.45303 | 0.00242 | -0.33556 | 0.00313 | -0.72119 | 0.00288 | -0.58245 |
| PLS | 0.01355 | -6.48721 | 0.01283 | -6.08951 | 0.01729 | -8.51291 | 0.01848 | -9.16937 |
| Lasso | 0.00199 | -0.09974 | 0.00182 | -0.00534 | 0.00204 | -0.12395 | 0.00181 | 0.00629 |
| ElasticNet | 0.00200 | -0.10585 | 0.00182 | -0.00330 | 0.00204 | -0.12382 | 0.00181 | 0.00477 |
| DL1 | 0.00179 | 0.01348 | 0.00178 | 0.01390 | 0.00178 | 0.01867 | 0.00178 | 0.01864 |
| DL2 | 0.00178 | 0.01376 | 0.00178 | 0.01433 | 0.00178 | 0.01922 | 0.00178 | 0.01850 |
| DL3 | 0.00179 | 0.01318 | 0.00179 | 0.01280 | 0.00179 | 0.01710 | 0.00178 | 0.01792 |
| DL4 | 0.00186 | -0.02858 | 0.00187 | -0.03178 | 0.00191 | -0.05172 | 0.00189 | -0.03882 |

Table 6: Prediction 1 month logarithm return of S&P 500. The training data is a cumulative window.

## A.2  3 month return prediction

| | Case 1 | | Case 2 | | Case 3 | | Case 4 | |
|---|---|---|---|---|---|---|---|---|
| | MSPE | $R^2_{OS}$ | MSPE | $R^2_{OS}$ | MSPE | $R^2_{OS}$ | MSPE | $R^2_{OS}$ |
| OLS | 0.00767 | -0.27961 | 0.00839 | -0.39954 | 0.00802 | -0.33652 | 0.00871 | -0.44996 |
| Ridge | 0.00670 | -0.11741 | 0.00679 | -0.13296 | 0.00693 | -0.15497 | 0.00694 | -0.15661 |
| PLS | 0.00762 | -0.27214 | 0.00821 | -0.37061 | 0.00795 | -0.32459 | 0.00836 | -0.39237 |
| Lasso | 0.00602 | -0.00373 | 0.00602 | -0.00457 | 0.00605 | -0.00696 | 0.00604 | -0.00648 |
| ElasticNet | 0.00602 | -0.00513 | 0.00602 | -0.00455 | 0.00605 | -0.00692 | 0.00604 | -0.00659 |
| DL1 | 0.00593 | 0.01045 | 0.00594 | 0.00927 | 0.00595 | 0.00915 | 0.00595 | 0.00937 |
| DL2 | 0.00593 | 0.01023 | 0.00593 | 0.01099 | 0.00595 | 0.00970 | 0.00593 | 0.01266 |
| DL3 | 0.00593 | 0.01074 | 0.00593 | 0.01042 | 0.00593 | 0.01233 | 0.00595 | 0.00911 |
| DL4 | 0.00600 | -0.00084 | 0.00600 | -0.00041 | 0.00607 | -0.01173 | 0.00595 | 0.00889 |

Table 7: Prediction 3 month logarithm return of S&P 500. The training data is a fixed length 600 month moving window.

| | Case 1 | | Case 2 | | Case 3 | | Case 4 | |
|---|---|---|---|---|---|---|---|---|
| | MSPE | $R^2_{OS}$ | MSPE | $R^2_{OS}$ | MSPE | $R^2_{OS}$ | MSPE | $R^2_{OS}$ |
| OLS | 0.06657 | -10.09405 | 0.08616 | -13.35903 | 0.05972 | -8.96527 | 0.10663 | -16.79488 |
| Ridge | 0.01306 | -1.17639 | 0.01444 | -1.40593 | 0.01343 | -1.24067 | 0.01482 | -1.47293 |
| PLS | 0.06816 | -10.35922 | 0.09308 | -14.51277 | 0.08010 | -12.36732 | 0.09960 | -15.62057 |
| Lasso | 0.00798 | -0.32950 | 0.00756 | -0.26015 | 0.00719 | -0.19937 | 0.00714 | -0.19080 |
| ElasticNet | 0.00797 | -0.32892 | 0.00762 | -0.27018 | 0.00729 | -0.21706 | 0.00735 | -0.22640 |
| DL1 | 0.00571 | 0.04783 | 0.00571 | 0.04767 | 0.00571 | 0.04737 | 0.00571 | 0.04680 |
| DL2 | 0.00571 | 0.04760 | 0.00571 | 0.04759 | 0.00570 | 0.04799 | 0.00570 | 0.04799 |
| DL3 | 0.00572 | 0.04680 | 0.00572 | 0.04678 | 0.00571 | 0.04655 | 0.00571 | 0.04647 |
| DL4 | 0.00602 | -0.00360 | 0.00603 | -0.00409 | 0.00623 | -0.03891 | 0.00618 | -0.03126 |

Table 8: Prediction 3 month logarithm return of S&P 500. The training data is a cumulative window.

## A.3  12 month return prediction

| | Case 1 | | Case 2 | | Case 3 | | Case 4 | |
|---|---|---|---|---|---|---|---|---|
| | MSPE | $R^2_{OS}$ | MSPE | $R^2_{OS}$ | MSPE | $R^2_{OS}$ | MSPE | $R^2_{OS}$ |
| OLS | 0.02903 | -0.12388 | 0.03320 | -0.28541 | 0.03308 | -0.28480 | 0.03700 | -0.43717 |
| Ridge | 0.02788 | -0.07920 | 0.02864 | -0.10883 | 0.02917 | -0.13287 | 0.02991 | -0.16163 |
| PLS | 0.02894 | -0.12034 | 0.03233 | -0.25154 | 0.03018 | -0.17200 | 0.03322 | -0.29019 |
| Lasso | 0.02703 | -0.04662 | 0.02700 | -0.04529 | 0.02709 | -0.05211 | 0.02706 | -0.05107 |
| ElasticNet | 0.02714 | -0.05069 | 0.02711 | -0.04954 | 0.02713 | -0.05371 | 0.02718 | -0.05553 |
| DL1 | 0.02496 | 0.03361 | 0.02497 | 0.03324 | 0.02504 | 0.02756 | 0.02502 | 0.02824 |
| DL2 | 0.02494 | 0.03450 | 0.02496 | 0.03356 | 0.02499 | 0.02948 | 0.02496 | 0.03060 |
| DL3 | 0.02499 | 0.03268 | 0.02499 | 0.03272 | 0.02516 | 0.02296 | 0.02505 | 0.02729 |
| DL4 | 0.02607 | -0.00936 | 0.02584 | -0.00041 | 0.02664 | -0.03445 | 0.02628 | -0.02076 |

Table 9: Prediction 12 month logarithm return of S&P 500. The training data is a fixed length 600 month moving window.

| | Case 1 | | Case 2 | | Case 3 | | Case 4 | |
|---|---|---|---|---|---|---|---|---|
| | MSPE | $R^2_{OS}$ | MSPE | $R^2_{OS}$ | MSPE | $R^2_{OS}$ | MSPE | $R^2_{OS}$ |
| OLS | 0.47498 | -14.73602 | 0.51265 | -15.98412 | 0.65768 | -21.26662 | 0.69387 | -22.49201 |
| Ridge | 0.09795 | -2.24502 | 0.10592 | -2.50926 | 0.10892 | -2.68761 | 0.11665 | -2.94934 |
| PLS | 0.46656 | -14.45711 | 0.47528 | -14.74595 | 0.56927 | -18.27361 | 0.56701 | -18.19688 |
| Lasso | 0.04705 | -0.55869 | 0.04714 | -0.56159 | 0.04596 | -0.55601 | 0.04433 | -0.50088 |
| ElasticNet | 0.04634 | -0.53519 | 0.04605 | -0.52564 | 0.04480 | -0.51667 | 0.04258 | -0.44144 |
| DL1 | 0.02619 | 0.13242 | 0.02608 | 0.13601 | 0.02625 | 0.11129 | 0.02611 | 0.11611 |
| DL2 | 0.02617 | 0.13290 | 0.02611 | 0.13505 | 0.02618 | 0.11352 | 0.02614 | 0.11492 |
| DL3 | 0.02624 | 0.13071 | 0.02605 | 0.13687 | 0.02634 | 0.10838 | 0.02614 | 0.11493 |
| DL4 | 0.03016 | 0.00069 | 0.03019 | -0.00017 | 0.03355 | -0.13583 | 0.03240 | -0.09704 |

Table 10: Prediction 12 month logarithm return of S&P 500. The training data is a cumulative window.

# B   Dropout

Dropout Hinton and Salakhutdinov (2006) and Srivastava et al. (2014) is a technique designed to avoid over-fitting in the training process. Input dimensions in $X$ are removed randomly with a given probability $p$. This affects the underlying loss function and optimization problem. For example, if $\mathcal{L}(Y, \hat{Y}) = \|Y - \hat{Y}\|_2^2$, where $\hat{Y} = WX$. When marginalizing over the randomness, we have a new objective

$$\arg\min_W \mathbb{E}_{D \sim \text{Ber}(p)} \|Y - W(D \star X)\|_2^2,$$

which is equivalent to

$$\arg\min_W \|Y - pWX\|_2^2 + p(1-p)\|(\text{diag}(X^\top X))^{\frac{1}{2}} W\|_2^2,$$

which is ridge a penalty under a $g$-prior. The dropout architecture is

$$\tilde{Y}_i^{(l)} = D^{(l)} \star X^{(l)}, \quad \text{where } D^{(l)} = (D_1^{(l)}, \cdots, D_p^{(l)}) \sim \text{Ber}(p)$$

$$Y_i^{(l)} = f(W_i^{(l)} X^{(l)} + b_i^{(l)})$$

where in effect, the input $X$ is replaced by $D \star X$, where $\star$ denotes the element-wise product and $D$ is a matrix of independent Bernoulli Ber$(p)$ distributed random variables.

## C   Deep Long-Short-Term-Memory

Traditional recurrent neural nets (RNNs) can learn complex temporal dynamics via the set of deep recurrence equations

$$Z_t = f(W_{xz}X_t + W_{zz} + b_x),$$

$$Y_t = f(W_{hz}Z_t + b_z),$$

where $Y_t$ is the output at time $t$, $X_t$ is the input, $Z_t$ is the hidden layer with N hidden units. For length $T$ the updates are computed sequentially.

Long-short-term-memories (LSTMs) are a particular form of recurrent network which provide a solution by incorporating memory units, see Hochreiter and Schmidhuber (1997). This allows the network to learn when to forget previous hidden states and when to update hidden states given new information. Models with hidden units with varying connections within the memory unit have been proposed in the literature with great empirical success. Specifically, in addition to a hidden unit, LSTMs include an input gate, a forget gate, an input modulation gate, and a memory cell. The memory cell unit combines the previous memory cell unit which is modulated by the forget and input modulation gate together with the previous hidden state, modulated by the input gate. These additional cells enable an LSTM architecture to learn extremely complex long-term temporal dynamics that a vanilla RNN is not capable of. Additional depth can be added to LSTMs by stacking them on top of each other, using the hidden state of the LSTM as the input to the next layer.

$$\text{State} \quad h_t = \sigma(o_t) \odot \tanh(c_t)$$

$$\text{Equations} \quad c_t = \sigma(f_t) \odot c_{t-1} + \sigma(i_t) \odot \tanh(k_t)$$

$$\begin{pmatrix} i_t \\ k_t \\ f_t \\ o_t \end{pmatrix} = \begin{pmatrix} W_{ix} & W_{ih} \\ W_{kx} & W_{kh} \\ W_{fx} & W_{fh} \\ W_{ox} & W_{oh} \end{pmatrix} \begin{pmatrix} x_t \\ h_{t-1} \end{pmatrix} + \begin{pmatrix} b_i \\ b_k \\ b_f \\ b_o \end{pmatrix} \tag{5}$$

where $\odot$ denotes element-wise vector product. The term $\sigma(f_t) \odot c_{t-1}$ introduces the long-range dependence. $k_t$ is new information flow to the current cell. The states $(f_t, i_t)$ controls weights of past memory and new information. $f_t$ is also called forget gate. The parameters $(W, b)$ of the stacked weight and bias vectors are learned by stochastic gradient descend (SGD) in `TensorFlow`. LSTM cell is defined by a state equation which is updated deterministically as

$$\begin{pmatrix} h_t \\ c_t \end{pmatrix} = \text{LSTMCell}(y_t, h_{t-1}) \tag{6}$$

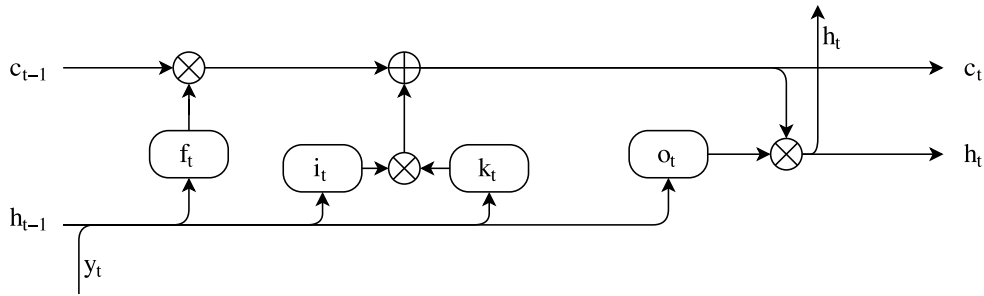

Figure 5: Structure of an LSTM cell

Figure 5 demonstrates the architecture of one LSTM cell. Let $y_t$ denote the observed time series and $h_t$ a hidden state. $c_t$ is the "memory" pass through multiple LSTM cells. The hidden state $h_t$ is generated using another hidden cell state, $c_t$ that will be generated so long term dependencies are allowed to flow in the network. The output state, $h_t$ is generated by a sequence of transformations known as an LSTMCell operator.

The key addition, compared to an RNN, is the hidden state $c_t$, the information is added or removed from the memory state via layers defined via a sigmoid function $\sigma(x) = (1 + e^{-x})^{-1}$ and point-wise multiplication $\otimes$. The first gate $f_t \otimes c_{t-1}$, called the forget gate, allows to throw away some data from the previous cell state. The next gate, $i_t \otimes k_t$, called the input gate, decides which values will be updated. Then the new cell state is a sum of the previous cell state, passed through the forgot gate selected components. This provides a mechanism for dropping irrelevant information from the past and adding relevant information from the current time step. Finally, the output layer, $o_t \otimes \tanh(c_t)$, returns *tanh* applied to the hidden state with some of the entries removed.